%% file: main.tex
\documentclass[conference]{IEEEtran}
\IEEEoverridecommandlockouts
\usepackage{cite}
\usepackage{amsmath,amssymb,amsfonts}
\usepackage{xcolor}
\usepackage[ruled,lined]{algorithm2e}
\usepackage{graphicx}
\usepackage{multirow}
\usepackage{hyperref}
\usepackage{subcaption}

\usepackage{tabularx}
\usepackage{booktabs}
\usepackage{epstopdf}
\usepackage{pifont}
\usepackage{color}
\usepackage{svg}
\usepackage{colortbl}

\graphicspath{{figs/}}

\usepackage{textcomp}
\usepackage{xcolor}
\def\BibTeX{{\rm B\kern-.05em{\sc i\kern-.025em b}\kern-.08em
    T\kern-.1667em\lower.7ex\hbox{E}\kern-.125emX}}

\input{macro}

\begin{document}
\newcommand{\xyli}[1]{\textcolor{blue}{#1}}

\title{\topic: \underline{K}inematic-Phrases-\underline{E}nhanced \underline{T}ext-to-Motion Generation via Fine-grained \underline{A}lignment}

\author{
    \IEEEauthorblockN{Yu Jiang\textsuperscript{*}}
    \IEEEauthorblockA{\textit{Zhiyuan College} \\
    \textit{Shanghai Jiao Tong University}\\
    Shanghai, China \\
    jy\_15924374500@sjtu.edu.cn}
    \and
    \IEEEauthorblockN{Yixing Chen\textsuperscript{*}}
    \IEEEauthorblockA{\textit{Zhiyuan College} \\
    \textit{Shanghai Jiao Tong University}\\
    Shanghai, China \\
    polaris\_dane@sjtu.edu.cn}
    \and
    \IEEEauthorblockN{Xingyang Li\textsuperscript{*}}
    \IEEEauthorblockA{\textit{Zhiyuan College} \\
    \textit{Shanghai Jiao Tong University}\\
    Shanghai, China \\
    brucelee\_sjtu@sjtu.edu.cn}
}

\maketitle

\footnotetext[1]{\textit{* Equal Contributions}}

\begin{abstract}
\input{abstract}
\end{abstract}


\input{intro}

\input{bg}
\input{method}

\input{eval}
\input{conclusion}
\input{acknowledgement}

\bibliographystyle{ieeetr}
\bibliography{references}

\end{document}

%% file: macro.tex
\ifx\figurename\undefined \def\figurename{Figure}\fi
\renewcommand{\figurename}{Fig.}
\renewcommand{\paragraph}[1]{\textbf{#1} }

\newcommand{\topic}{\textsc{KETA}\xspace}

\newcommand{\RNum}[1]{\uppercase\expandafter{\romannumeral #1\relax}}


%% file: abstract.tex
Motion synthesis plays a vital role in various fields of artificial intelligence. 
Among the various conditions of motion generation, text can describe motion details elaborately and is easy to acquire, making text-to-motion(T2M) generation important. State-of-the-art T2M techniques mainly leverage diffusion models to generate motions with text prompts as guidance, tackling the many-to-many nature of T2M tasks.
However, existing T2M approaches face challenges, given the gap between the natural language domain and the physical domain, making it difficult to generate motions fully consistent with the texts. 

We leverage kinematic phrases(KP), an intermediate representation that bridges these two modalities, to solve this. Our proposed method, \topic, decomposes the given text into several decomposed texts via a language model. It trains an aligner to align decomposed texts with the KP segments extracted from the generated motions. Thus, it's possible to restrict the behaviors for diffusion-based T2M models. \textit{During the training stage}, we deploy the text-KP alignment loss as an auxiliary goal to supervise the models. \textit{During the inference stage}, we refine our generated motions for multiple rounds in our decoder structure, where we compute the text-KP distance as the guidance signal in each new round. Experiments demonstrate that \topic achieves up to 1.19$\times$, 2.34$\times$ better R precision and FID value on both backbones of the base model, motion diffusion model. Compared to a wide range of T2M generation models. \topic achieves either the best or the second-best performance.

%% file: intro.tex
\section{Introduction}

Motion synthesis is an important task that is widely applied in today's artificial intelligence technologies~\cite{zhu2023human}, enhancing the development of the interaction between humans and computers~\cite{liu2024control, jiang2024scaling}, computer vision~\cite{roditakis2024towards, gong2024demosdynamicenvironmentmotion}, et cetera. Substantial guidance information is required to describe motions, making text-to-motion(T2M) generation essential. Currently, many T2M models~\cite{ahn2018text2action, tevet2022motionclip, athanasiou2022teach} exploit traditional machine-learning techniques to generate photo-realistic motions. Among these methods, diffusion-based T2M models~\cite{zhang2022motiondiffuse, chen2023executing} offer the best performance by tackling the many-to-many nature of T2M tasks, where the natural language yields ambiguity and the motion has various descriptions.

\begin{figure*}[!tp]
    \centering
    \includegraphics[width=\linewidth]{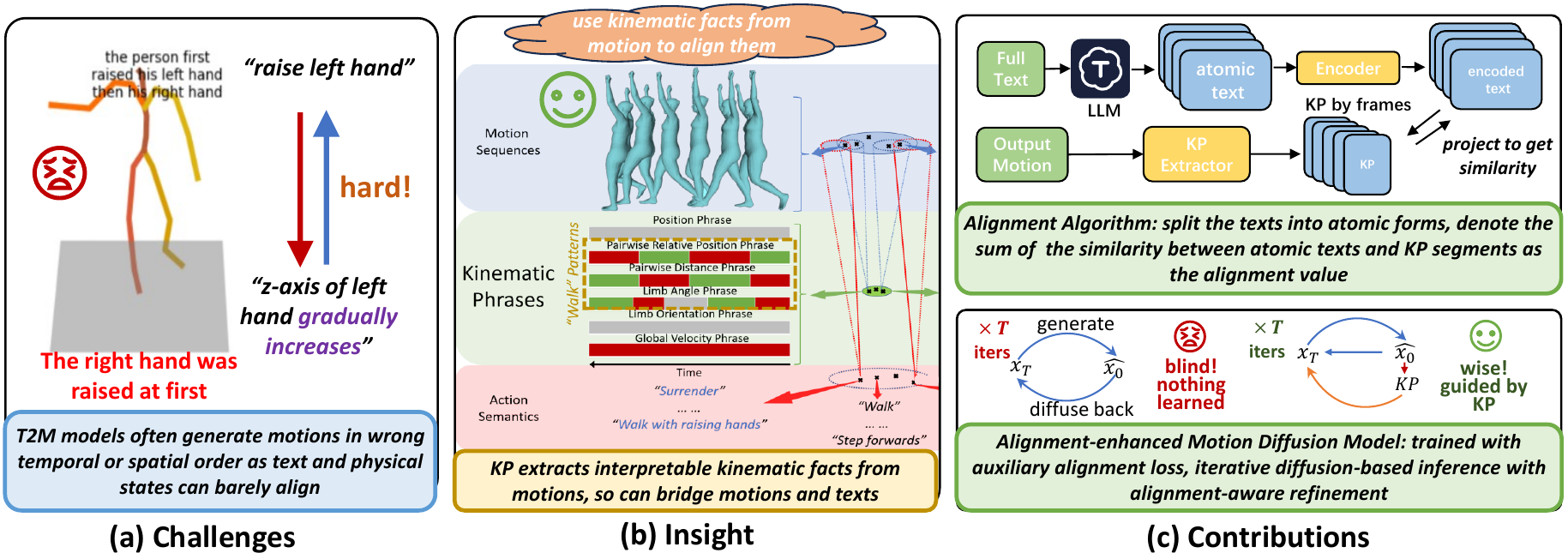}
    \caption{Overview of {\topic}, a physical-state-aware T2M model via fine-grained alignment between text and KP.}
    \label{fig:overview}
\end{figure*}

However, another problem is the intrinsic gap between language and motion space. Language is abstract and concise for better human usage and communication, making the corresponding space sparse and discrete. Motion is continuous as it is directly based on physical space. Such a gap results in the degradation of motion quality. Previous T2M models fail to cope with texts with spatial and temporal objectives, generating either motion unrelated to the text or wrong in temporal order and spatial position, as illustrated in Fig.~\ref{fig:overview}(a).

Researchers propose Kinematic Phrases (KP)~\cite{liu2025bridging} to extract abstract and general kinematic facts from motions. As illustrated in Fig.~\ref{fig:overview}(b), KP captures the innate features of motion by computing relative position and its change over time. KP appears in comprehensible natural language and sticks to objective kinematic facts.

Inspired by KP, we leverage it to construct fine-grained alignments to restrict the physical behaviors of motions. Our key insight is that \textbf{(1)KP bridges motion and action-level knowledge about the motions, possibly enhancing the T2M process, and (2)sequential fine-grained sentences provide detailed information on action semantics and the temporal relationship among the motions.}
Specifically, we first introduce an LLM agent to split the original text into decomposed texts, making a finer text granularity. Then, we extract KP from the dataset, acquiring their spatial and temporal features and corresponding constraints. We use an encoder to project text information onto the KP space. Next, our proposed domain model identifies a Gaussian distribution weight for the particular segment of KP assigned to the decomposed text. Our alignment goal is to minimize the distance on the KP space by co-training the encoder and the domain model.

Based on the fine-grained alignment approach, we propose \topic to enhance the training and inference process of diffusion-based T2M models. In the training stage, we add an auxiliary alignment loss for text and KP extracted from generated motions. Moreover, we smooth the KP with the $\text{tanh}$ operator to allow back propagation instead of solely considering its sign. We also provide a decoder model, utilizing the difference between the generated KP and text as an additional token with a mask, and train the model to predict the clean motion with this guidance signal. Consequently, the model can step towards more physically consistent motion generation.

During the inference stage, we expand the diffusion process into $T$ cycles of full denoising and diffusing. We utilize our text-KP alignment results in this process by constantly using the difference between the current generated KP and the given text as a guidance signal for denoising in the next step. This allows us to generate motion in a closed-loop manner, continuously refining the motion generated with guidance toward the given text for pre-defined steps.

Our contributions can be summarized as follows:

\begin{itemize}
    \item We prompt a language model to split the texts into decomposed parts and introduce a method to align them by training an encoder to project the decomposed texts onto the KP space and evaluate their similarities with the corresponding KP segments.
    \item 
    We use the fine-grained text-KP alignment as an auxiliary goal and supervision for training. We augment the KP with smoothing to allow back propagation in training. 
    \item 
    During the inference stage of the diffusion-based T2M model, we use a guidance signal based on the relative distance between the current generated motion's KP and text on the aligned space to refine the current motion in each round of iteration towards our expected goal in a closed-loop manner.
    \item 
    The encoder model of \topic gains up to 1.19$\times$, 2.34$\times$ better R-precision and FID value than both backbones of our base motion diffusion model and is either the best or the second-best model compared with various T2M models on these metrics.
\end{itemize}

%% file: bg.tex
\section{Background and Related Work}
\label{sec:bg}
\subsection{Motion Diffusion Model}

The Human Motion Diffusion Model(MDM)~\cite{zhang2022motiondiffuse} is a classifier-free diffusion transformer-based model for motions corresponding to different prompts, like texts and actions.

\paragraph{Diffusion Models.}
We begin by explaining the theoretical basis of diffusion models~\cite{ho2020denoising, song2020denoising}. Diffusion models train a model for predicting the noise, thus generating results. By modeling the process of diffusing data in a certain distribution to pure noise and the reverse process of denoising pure noise back to in-distribution data. In the diffuse process~\cite{ho2020denoising}, Gaussian noise is iteratively added according to a given schedule denoted $\beta_{1:T}$.

$$
q(x_t| x_{t-1})=\mathcal{N}(x_t;\sqrt{1-\beta_t}x_{t-1}, \beta_t\mathbf{I})
$$

Let $\alpha_t=1-\beta_t,\bar{\alpha_t}=\alpha_1\alpha_2\cdots\alpha_t$, then we can directly obtain $x_t$ by sampling only once. The noise here follows a normal distribution: $\epsilon\sim\mathcal{N}(0,\mathbf{I})$.

$$
x_t=\sqrt{\bar{\alpha_t}}x_{0}+\sqrt{1-\bar{\alpha_t}}\epsilon
$$

For the reverse direction, we try to denoise sampled noise to the original distribution by the following formula, which can be derived from Bayes' law $q(x_{t-1}| x_t,x_0)=\dfrac {q(x_{t}| q_{x_{t-1}})q(x_{t-1}| x_0)} {q(x_t| x_0)}$.

$$
p_{\theta}(x_{t-1}| x_t)=\mathcal{N}(x_{t-1};\mu_{\theta}(x_t,t),\Sigma_{\theta}(x_t,t))
$$

Note that the noise $\epsilon$ is unknown at this point, so we utilize a model to predict the noise added with the input of the current diffuse result $x_t$ and the diffuse timestep $t$.

At training time, we diffuse $x_0$ to random timestep $t$ with sampled noise $\epsilon$ and utilize the model to predict the noise under the loss function of:

$$
\mathcal{L}=\mathbb{E}_{x_0,t}[||\epsilon-\epsilon_{\theta}(\sqrt{\bar{\alpha_t}}x_0+\sqrt{1-\bar{\alpha_t}}\epsilon, t)||^2]
$$

\paragraph{Classifier-free Guidance.}
Classifier-free guidance~\cite{ho2022classifier} is a technique for diffusion models that improves the sample quality without relying on an external classifier. Classifier-free guidance diffusion models train an unconditional denoising diffusion model $p_\theta(z)$ alongside a conditional model, $p_\theta(z|c)$, both parameterized by a single neural network. A null token is deemed the input as the class identifier $c$ for the unconditional model.

During the training phase, we randomly choose $c$ with probability $p_\text{uncond}$, improving joint training of both models while eliminating the need for extra model complexity of parameters. During the sampling phase, classifier-free guidance is implemented by adding the conditional and unconditional scores together linearly through a score estimation formula:

$$
\tilde{s}(z|c) = (1 + w) s(z|c) - w s(z)
$$

Here, $w$ denotes the strength of the guidance, while $s(z|c), s(z)$ denotes the score estimates from the conditional and unconditional models, respectively. The added score guides the sampling process, achieving the trade-off in sample quality without relying on classifier gradients. Classifier-free guidance diffusion models have demonstrated superior samples over vanilla sampling techniques and are widely deployed in SOTA diffusion-based models~\cite{10377858, yang2024cogvideox}.

\paragraph{Motion Diffusion Model.}
MDM utilizes a CLIP model to obtain text embedding, which, along with timestep $t$, are projected into a token $z_{tk}$. A transformer encoder is then deployed to predict the final clean motion from the concatenation of noisy motion sequence and the input oken $z_{tk}$.

\begin{figure}[t]
    \centering
    \includegraphics[width=\columnwidth]{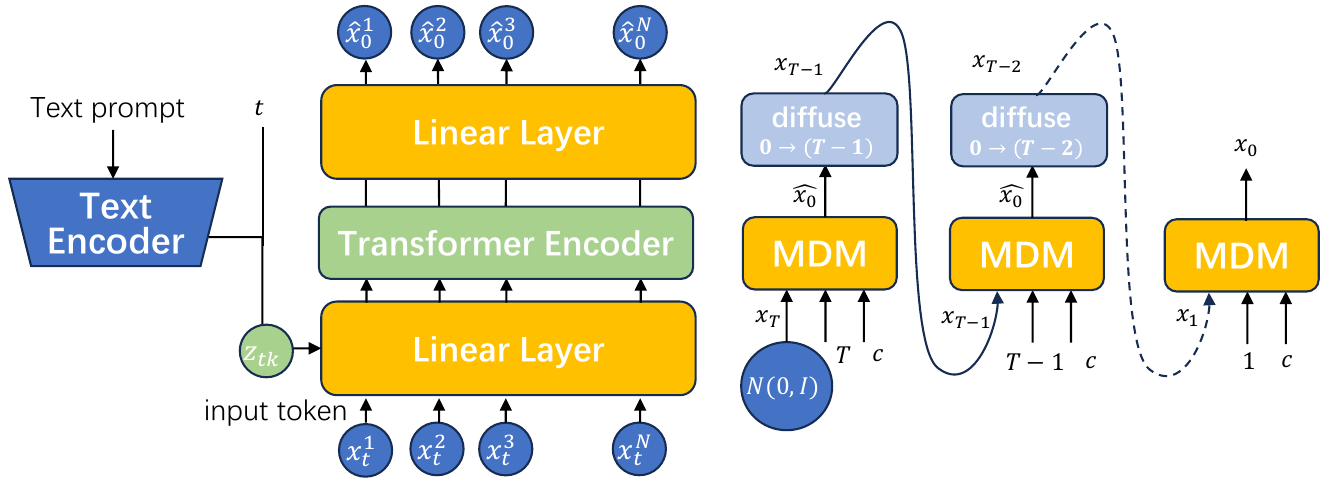}
    \caption{\textbf{Left: The Motion Diffusion Model (MDM) overview.} The input is the motion sequence of length, the current timestep $t$, and a conditioning code, then a Text Encoder projects and adds them to get the input token $z_{tk}$. During each step, the clean motion is predicted. \textbf{Right: MDM Sampling.} Given a condition $c$, the model iterates to get the refined motion predicting a clean sample and diffuses it back in each round.}
    \label{fig:mdm}
\end{figure}

Fig.~\ref{fig:mdm} illustrates the inference stage of MDM. At each timestep $t$, MDM predicts the clean motion from the previous diffused result $x_t$ and then diffuses it back to $x_{t-1}$, which can be considered refining generated motion.

\subsection{KP}
Motion understanding is an important task that is widely applied in autonomous settings. However, assessing the consistency between action semantics and motions is challenging because motions usually convey multiple intentions in different contexts, and various motions can depict action semantics.

Kinematic Phrases(KP) are proposed to resolve the gap between motion and action semantics by focusing on objective kinematic facts, like the movements of different joints or organs. Also, KP uses a signed bit to denote the direction of these movements. Consequently, the change in the sign of KP represents the change in the person's physical condition. Through such abstract and interpretable representation, we can leverage KP to learn objective physical information about the motions, thus adding proper restrictions to motion generation to align motions with kinematic behaviors.

%% file: method.tex

\section{Methods}

\subsection{Overview}
\label{sec:methods:overview}

T2M models aim to synthesize human motion from a given text. \topic contains a text decomposition agent(DA), an alignment model(AA), and an MDM-based motion generation model. 
When a text description of the wanted motion is provided, it is first decomposed into several sentences by the DA, which are subsequently encoded by the Llama~\cite{dubey2024llama} language model into text embeddings. The AA then computes each decomposed text's domain range and weight; thus, it can align the text to the corresponding domain on the KP space.

Our vanilla approach is to directly add the align loss computed by the AA from the whole sentence into the aggregate loss of the MDM by using the encoder backbone. Therefore, we can supervise the motion generation process from a physically coherent perspective. To further enhance the model with the help of KP, we also adopt a decoder backbone for MDM with the decomposed texts as input and a guide token as guidance so that the inference process of MDM can be continuously refined. Specifically, we use the difference between the current generated motion's weighted KP and the text feature as the guide token to guide the refinement toward the desired description.


\subsection{Text Decomposition}

Existing methods for motion synthesis struggle to follow semantic instructions, especially when it comes to temporally and spatially specific instructions. Models must first comprehend these relations and then realize them in the physical space, making it hard to do it end-to-end. Instructions are needed to explicitly clarify these temporal and spatial relationships to tackle this. We believe that advanced large language models can play a critical role by decomposing the whole text into separate texts that follow the temporal and spatial order of the generated motions. Therefore, the motion generation model can better understand the inherent physical constraints.

\begin{figure*}[!tp]
    \centering
    \includegraphics[width=\linewidth]{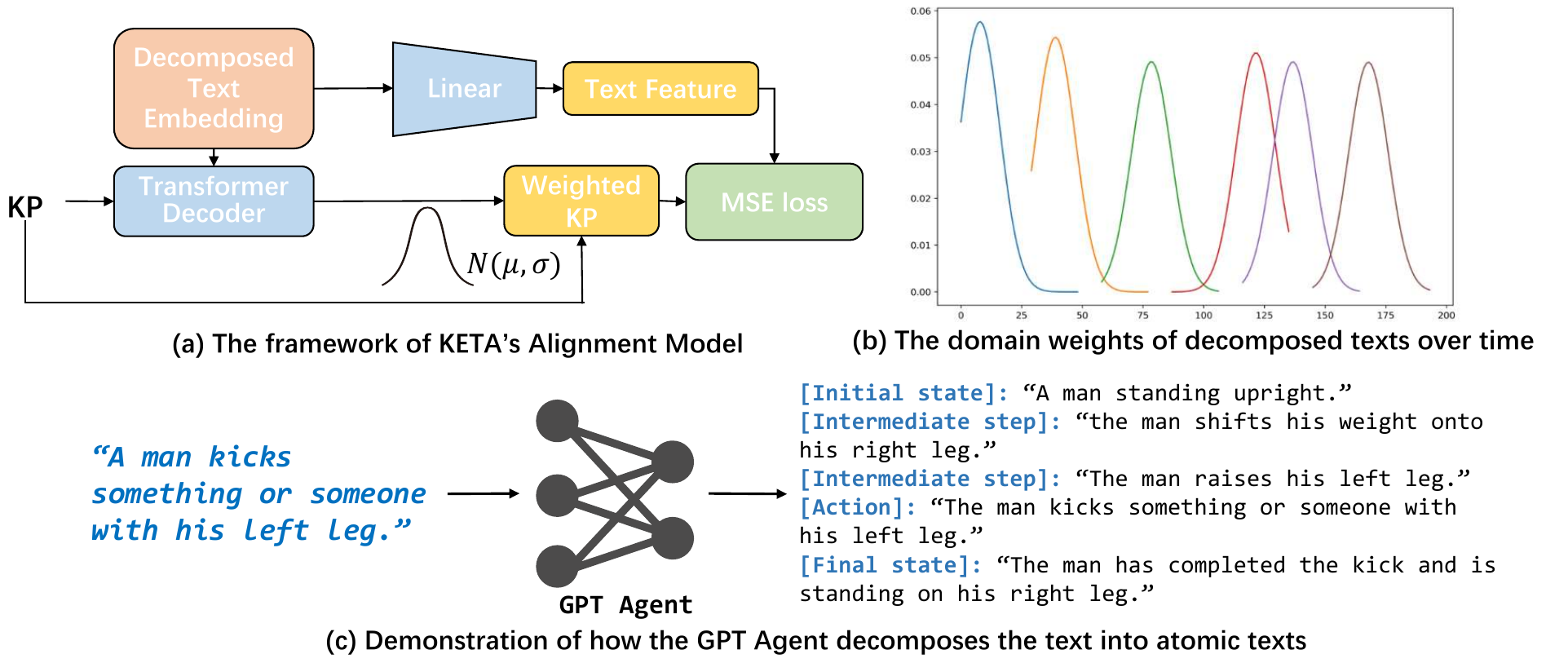}
    \caption{The text decomposition process and the alignment model structure.}
    \label{fig:align}
\end{figure*}

\topic uses an agent derived from GPT-4o-mini~\cite{gpt-4o} to decompose the given text with the following prompt. The agent generates decomposed texts ranging from 3 to at most 20 sentences for a given text prompt, having high fidelity in temporal order while reserving the text's original meaning. Fig.~\ref{fig:align} (c) depicts the whole process.

\begin{quote}
\textit{You are a highly specialized assistant designed to analyze and process textual descriptions of human actions. Your primary function is to decompose these descriptions into fine-grained actions arranged chronologically. Focus on detecting and interpreting sequence markers like 'then,' 'twice,' 'again,' and other words indicating repetitions or transitions. Ensure that your decomposition explicitly outlines: 1. The initial state of the posture or action. 2. Detailed intermediate steps. 3. The final state.}
\end{quote}

\subsection{Text-KP Alignment}

KP extraction is based on the human body joint sequence $X=\{x_i\mid x_i\in \mathbb{R}^{n_j\times 3}\}$ where $n_j$ is the number of joints, and $i$ is the current time frame. Several extraction functions $f_j$ compute values from the joint sequence. After that, the function values are passed through an indicator function for each frame, generating the final KP sequence of $\{-1,0,1\}^n$.

$$
KP_j(x_i)=\text{sign}(f_j(x_i))
$$

The original KP generation process has operations that do not have a gradient, such as the indicator function $\text{sign}$. To leverage KP in training, we replace them with smooth activation functions with available gradients.

$$
KP_j(x_i)=\tanh(f_j(x_i))
$$

The next step is aligning the KP sequence with the decomposed texts. As KP is different for every single frame, there's difficulty in aligning the whole KP sequence with some specific texts. As mentioned in~\ref{sec:methods:overview}, we compute a special weighted KP for a decomposed text and align the corresponding text feature with it.

Our insight is that each decomposed text takes effect in its time domain. This is intuitive because the first few decomposed texts correlate more with the first few frames of the KP sequence. Similarly, the motion at a given moment results from all decomposed texts with a domain covering that moment. Thus, we model the domain weight of each decomposed text as a Gaussian distribution, as shown in Fig.~\ref{fig:align} (b).

The reason for choosing such distribution is straightforward since we can divide the whole motion into a composition of three stages for each decomposed text, respectively: the transition from the previous motion segment, the current motion segment, and the transition to the next motion segment. Of the three stages, the second one is dominated by the current text only, while the other two result from intersection with different text. The current text's influence increases as the current action begins and fades away as the current action is done, matching the tendency of a Gaussian distribution.

We also provide a "feasible window" for each decomposed text to locate its distribution to ensure the overall temporal structure of the whole motion. Without the window, the weight distribution of different texts tends to converge to similar ones. For a sequence of motion of length $T$ with $n$ decomposed text, the range for the $i-$th window $[l_i, r_i]$ is defined by the following expression:

\[
\left\{
\begin{aligned}
l_i &= \frac{i}{n-1} \cdot \frac{T}{n} \cdot (n - 1 - \frac{1}{\log(n+2)}), \\
    r_i &= l_i + \frac{T}{n} \cdot \left(1 + \frac{1}{\log(n+2)}\right)
\end{aligned}
\right.
\]

A 2-layer MLP extracts text features of the same dimension as KP. As for the domain model, we use a transformer decoder architecture to decode the weight over the text domain with the text feature as the query. The weighted KP $\Omega_i$ is calculated by:

$$
\Omega_i=\sum\limits_{j\in\text{domain}_i}w_jKP_{j}
$$

In summary, the overall align loss for a decomposed text embedding sequence $\{T_i\}_{i=1}^{n}$ is:

$$
\mathcal{L}=\sum\limits_{i}||\Omega_i-\text{MLP}(T_i)||^2
$$

\subsection{Motion Generation with Iterative Refinement by Alignment}

Beyond controlling the motion generation process with physical constraints via adding auxiliary loss, we want to involve the decomposed texts directly in the diffusion model. However, the original implementation of MDM with a transformer encoder architecture fails to meet such need. We solve this by using the transformer decoder as the backbone, where the decomposed embeddings are concatenated as the query for cross-attention layers.


In the MDM framework, clean motion is directly generated for each timestep of denoising. It iteratively refines the generated motion by diffusing the generated clean motion to a decreasing timestep $t$ and denoise it again. However, the model fails to notice the problem in physical states with the current motion during this process, making the refinement rather blind. To resolve this, we utilize our alignment framework by introducing a guidance token into our model structure. 
By comparing the KP extracted from the current generated motion with the desired text description, we can inform the model of where refinement is needed and how to refine it. Moreover, we can consider the diffused motion in the training phase as motion that needs to be refined, which offers abundant data for training our module of guided refinement.

%% file: eval.tex
\section{Evaluation}

\subsection{Experimental Setup}

We briefly introduce the baseline models, datasets, and evaluation metrics we use to test our model's performance to help readers understand the experiment details holistically.

\paragraph{Baseline Models and Datasets.}
We evaluate the impact of \topic on the Human Motion Diffusion Model(MDM)~\cite{zhang2022motiondiffuse}, a diffusion-based motion generation model with both transformer encoder and decoder structure. We assess the model on the HumanML3D~\cite{guo2022humanml} dataset, which stems from the combination of the  HumanAct12~\cite{guo2020action2motion} and Amass~\cite{mahmood2019amass} and is considered a comprehensive dataset of human actions. Specifically, HumanML3D comprises 14646 motions and 44970 motion annotations. We also compare the performance of \topic with a wide range of state-of-the-art T2M models, including MDM(both encoder and decoder structure)~\cite{zhang2022motiondiffuse}, MAA~\cite{azadi2023make} and OMG~\cite{liang2024omg}.

\paragraph{Evaluation Metrics.}
We evaluate \topic-enhanced MDM via standard metrics introduced in a previous work~\cite{guo2022humanml}. The metrics involve Frechet Inception Distance (FID), diversity, and R-Precision. FID is a metric that evaluates the dissimilarity between the generated sample and the ground truth, which should be as low as possible. Diversity assesses the distribution range of motions, and the closer the diversity value between the assessed model and the ground truth, the better the model. R-Precision evaluates the similarity between text prompts and generated motions on a semantic level by searching the text features in the motions. This means a model is greater if it demonstrates a higher R-precision value.

\paragraph{Implementation.}
We implement our methodology of \topic on MDM's encoder and decoder backbone. For the encoder structure, we can only utilize the auxiliary alignment loss. For the decoder structure, the inference process is iteratively refined with alignment difference.
Our \topic-enhanced MDM is trained with $1500$ epochs, the same as the original MDM. The text encoder we use is a Llama 3.1 model, and we use GPT-4o-mini to decompose the text. We use a scaling coefficient $\lambda_{KP}$ to control the impact of text-to-KP alignment on the model. We set $\lambda_{KP}=0.0001$ in our implementation. We train our \topic-enhanced MDM on a single NVIDIA A100 GPU~\cite{nvidiaA100}. The full training process takes about 4 days.

\subsection{Performance}

\begin{table}[h]
\centering
\begin{tabular}{lccc}
\toprule
\textbf{Methods} & \textbf{R-Precision $\uparrow$} & \textbf{FID$\downarrow$} & \textbf{Diversity$\rightarrow$} \\
\midrule
Real & 0.797 & 0.002 & 9.503 \\
\midrule
MDM(encoder) & 0.611 & 0.544 & \cellcolor{yellow!20} \textbf{9.559} \\
MDM(decoder) & 0.621 & 0.567 & 9.425 \\
MAA & 0.675 & 0.774 & 8.230 \\
OMG & \cellcolor{green!20} \textbf{0.784} & 0.381 & 9.657 \\
\midrule
Ours(encoder) & 0.728 & \cellcolor{yellow!20} \textbf{0.279} & \cellcolor{green!20} \textbf{9.487} \\
Ours(decoder) & \cellcolor{yellow!20} \textbf{0.737} & \cellcolor{green!20} \textbf{0.242} & 9.605 \\
\bottomrule
\end{tabular}
\caption{The performance of \topic compared with various baselines.}
\label{tab:perf}
\end{table}

\begin{figure}[t]
    \centering
    \begin{subfigure}[b]{0.45\columnwidth}
        \includegraphics[width=\linewidth]{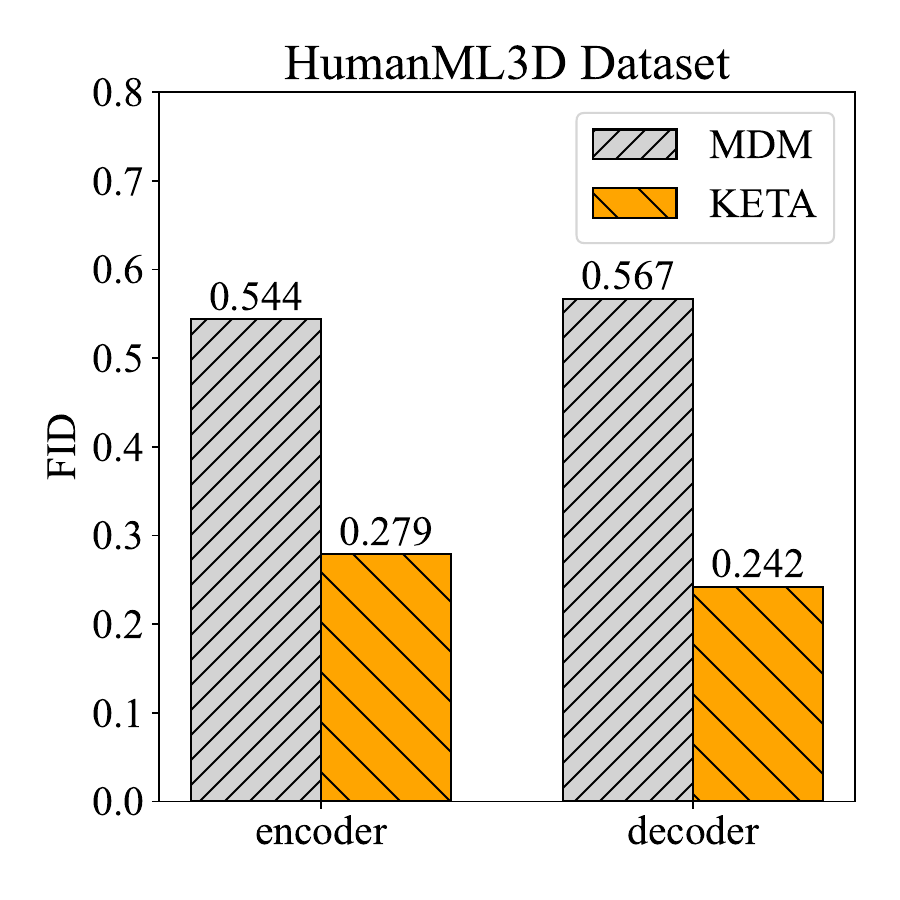}
        \caption{\topic and MDM's comparison in FID value}
        \label{fig:fid}
    \end{subfigure}
    \hfill
    \begin{subfigure}[b]{0.45\columnwidth}
        \includegraphics[width=\linewidth]{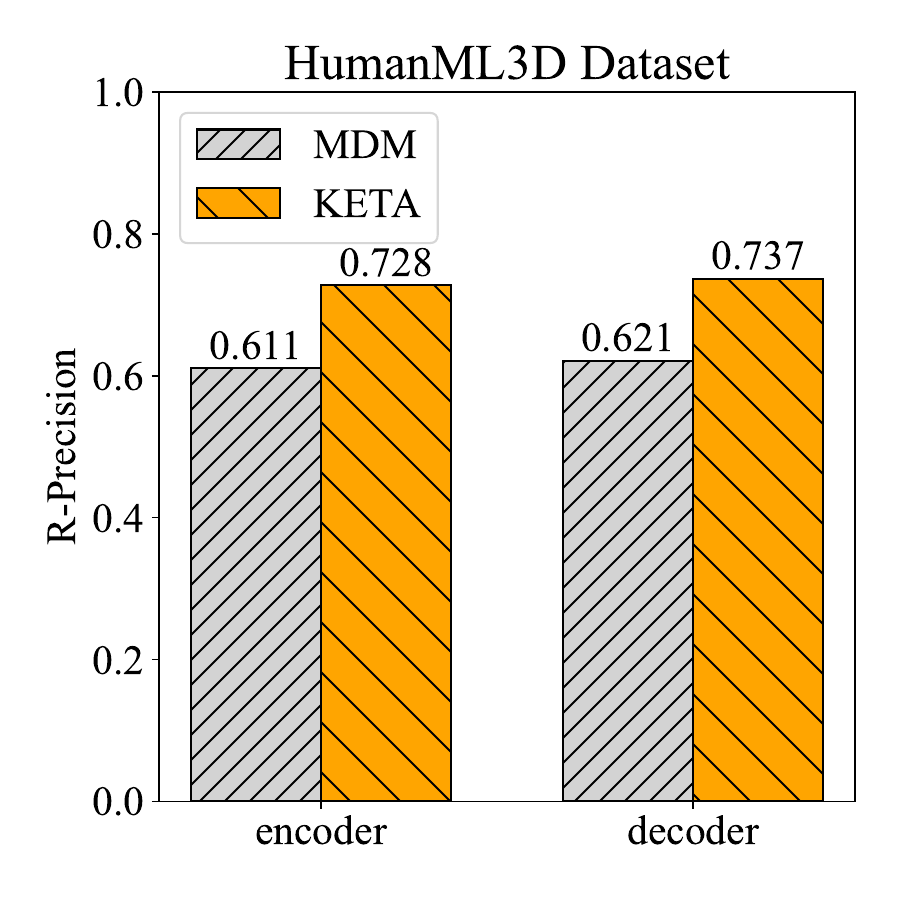}
        \caption{\topic and MDM's comparison in R-precision}
        \label{fig:r}
    \end{subfigure}
    \caption{Comparison between \topic and MDM, both the transformer encoder and decoder structure of \topic outperform their MDM counterparts in both FID and R-precision.}
\label{fig:cmp}
\end{figure}

\begin{figure*}[!tp]
    \centering
    \includegraphics[width=\linewidth]{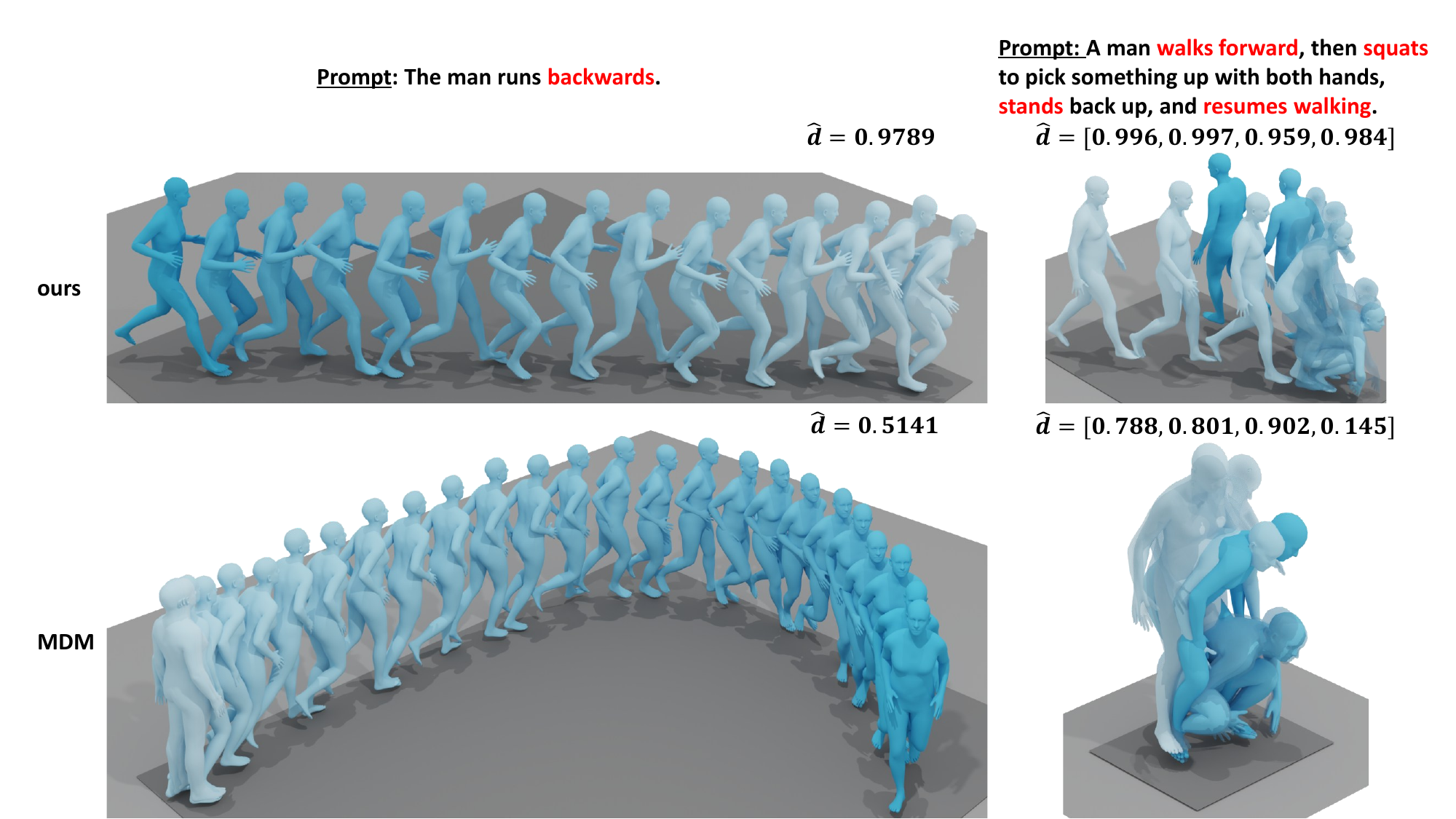}
    \caption{The motions generated by \topic compared with MDM, our generated motions are consistent with the text prompts, showing higher text-to-motion similarity. Meanwhile, MDM fails to achieve spatial and temporal consistency.}
    \label{fig:results}
\end{figure*}

Tab.~\ref{tab:perf} demonstrates the performance of \topic-enhanced MDM compared with multiple SOTA T2M models on various metrics. Specifically, illustrated in Fig.~\ref{fig:cmp}, compared with the original MDM model, the encoder backbone of \topic achieves 1.19$\times$ higher R-Precision value, 1.99$\times$ lower FID value. In comparison, the decoder backbone of \topic achieves 1.18$\times$ higher R-precision value and $2.34\times$ lower FID value. Both backbones surpass the base model in all aspects. Moreover, compared with the SOTA T2M models, \topic achieves the best result in both FID and Diversity. \topic is also second-best in R-Precison.

\subsection{Impact of our Fine-grained Alignment}

Fig.~\ref{fig:results} depicts the motion generated by \topic and the base model MDM. The deepening colors represent the passage of time, with darker hues corresponding to later states. For the prompt, "A man runs backward," the motion generated by MDM illustrates a person running forward, violating the prompt. On the other hand, \topic-enhanced MDM perfectly adheres to the prompt, suiting spatial constraints. Regarding the prompt, "A man walks forward, then squats to pick something up with both hands, stands back up, and resumes walking," the motion generated by MDM doesn't walk in the beginning or resume walking in the end. At the same time, our model suits this prompt, showing that our model resolves the temporal relationships. 

From a quantitative perspective, we also evaluate the cosine similarity between the text feature and the motions generated by our model, \topic, via the alignment model we designed. For the first prompt, the overall cosine similarity $\hat{d}$ between the text feature and the motions generated by \topic is higher than that in MDM. For the second prompt, the MDM fails to be consistent with the first and the last action, so it has a substantially lower text-to-motion similarity for the first and the last decomposed texts. However, the motions generated by our model \topic demonstrated high capability in aligning with all the decomposed texts, showing the strong power of our KP-enhanced T2M generation via fine-grained alignment.

\subsection{Ablation Study}

We test the performance of \topic with the fine-grained alignment and the alignment between the full KP and the full text. As shown in Tab.~\ref{tab:ablation}, although the full align encoder backbone of \topic has slightly better FID, it has worse R-precision compared with its fine-grained align counterpart, meaning the fine-grained align model has better ability in aligning with the prompt text. Moreover, the fine-grained align model has a smaller difference in diversity to ground truth, meaning they are more similar. In conclusion, our fine-grained alignment approach generates temporally and spatially correct motions better than directly employing full-text alignment.

\begin{table}[h]
\centering
\begin{tabular}{lccc}
\toprule
\textbf{Methods} & \textbf{R-Precision $\uparrow$} & \textbf{FID$\downarrow$} & \textbf{Diversity$\rightarrow$} \\
\midrule
Real & 0.797 & 0.002 & 9.503 \\
\midrule
KETA enc.(fine-grained align) & \cellcolor{green!20} \textbf{0.728} & 0.279 & \cellcolor{green!20} \textbf{9.487} \\
KETA enc.(full align) & 0.707 & \cellcolor{green!20} \textbf{0.264} & 9.536 \\
\bottomrule
\end{tabular}
\caption{Comparison between the performance of \topic encoder structure with and without the fine-grained alignment.}
\label{tab:ablation}
\end{table}

%% file: conclusion.tex
\section{Conclusion}
In this paper, we propose \topic, a physical-behavior-aware text-to-motion generation with the help of fine-grained text-KP alignment, taking one step towards physical-consistent text-to-motion generation. \topic significantly optimizes both the encoder and the decoder backbone of the base motion diffusion model, achieving up to 1.19$\times$ and 2.34$\times$ better R-precision, FID value. Compared to a wide range of text-to-motion generation models. \topic achieves either the best or the second-best performance. We hope that in the future, more works can find inspiration in our work and focus on leveraging physical details to boost the quality of motion generation.

%% file: acknowledgement.tex
\section*{Acknowledgments}

We sincerely thank Xinpeng Liu, Qiaojun Yu, and Yonglu Li for providing insightful guidance on exploring the topic of T2M generation. We thank Yichao Zhong, Tianlang Zhao, Tongcheng Zhang, and Jinbo Hu for their valuable feedback on different stages of the work.